\documentclass[a4paper]{article}

\usepackage{INTERSPEECH2019}
\usepackage[utf8]{inputenc} 
\usepackage[T1]{fontenc}    
\usepackage{hyperref}       
\usepackage{url}            
\usepackage{booktabs}       
\usepackage{amsfonts}       
\usepackage{nicefrac}       
\usepackage{microtype}      
\usepackage{color}
\usepackage{graphicx}
\usepackage{amssymb}
\usepackage{cite}
\usepackage{enumitem}
\setlist[itemize]{leftmargin=*}

\title{Learning Problem-agnostic Speech Representations\\ from Multiple Self-supervised Tasks}
\name{Santiago Pascual$^1$, Mirco Ravanelli$^2$, Joan Serr\`{a}$^3$, 
      Antonio Bonafonte$^{1\ast}$\thanks{$^\ast$A.~Bonafonte is currently at Amazon Research, Cambridge, UK.}, 
      Yoshua Bengio$^{2,4}$}
\address{
  $^1$Universitat Polit\`{e}cnica de Catalunya, 
  $^2$Mila - Universit\'{e} de Montr\'{e}al, \\
  $^3$Telef\'{o}nica Research, 
  $^4$CIFAR Fellow}
\email{santi.pascual@upc.edu, mirco.ravanelli@gmail.com}

\makeatletter
\def\blfootnote{\xdef\@thefnmark{}\@footnotetext}
\makeatother

\begin{document}

\maketitle
\begin{abstract}
Learning good representations without supervision is still an open issue in machine learning, and is particularly challenging for speech signals, which are often characterized by long sequences with a complex hierarchical structure. Some recent works, however, have shown that it is possible to derive useful speech representations by employing a self-supervised encoder-discriminator approach. This paper proposes an improved self-supervised method, where a single neural encoder is followed by multiple workers that jointly solve different self-supervised tasks. The needed consensus across different tasks naturally imposes meaningful constraints to the encoder, contributing to discover general representations and to minimize the risk of learning superficial ones. Experiments show that the proposed approach can learn transferable, robust, and problem-agnostic features that carry on relevant information from the speech signal, such as speaker identity, phonemes, and even higher-level features such as emotional cues. In addition, a number of design choices make the encoder easily exportable, facilitating its direct usage or adaptation to different problems.
\end{abstract}

\section{Introduction}

The success of deep learning techniques strongly depends on the quality of the representations that are automatically discovered from data. 
These representations should capture intermediate concepts, features, or latent variables, and are commonly learned in a supervised way using large annotated corpora. Even though this is still the dominant paradigm, some crucial limitations arise. Collecting large amounts of annotated examples, for instance, is very costly and time-consuming.
Moreover, if not learned with a large pool of tasks~\cite{Serra18AIRD}, supervised representations are likely to be biased towards the considered problem, limiting their exportability to other problems and applications~\cite{transfer_learning}.

A natural way to mitigate these issues is unsupervised learning~\cite{Bengio_2012}. Unsupervised learning attempts to extract knowledge from unlabeled data, 
and can potentially discover representations that capture the underlying structure of such data. Several approaches have been proposed for unsupervised learning in the last decade. Notable examples are deep autoencoders~\cite{deep_autoencoder_1} and restricted Boltzmann machines~\cite{IEEEexample:rbm1}, which can be employed as a pre-training step for a subsequent supervised task like speech recognition \cite{IEEEexample:intro1}. More recent techniques include variational autoencoders~\cite{var_auto} and generative adversarial networks~\cite{gan}. 

A related sub-field that is gaining popularity, especially within the computer vision community, is self-supervised learning, where targets are computed from the signal itself~\cite{multi_task_self_sup,self_sup2,self_sup3}. This is often performed by applying known transforms or sampling strategies to the input data and using the resulting outcomes as targets. Some attempts have also been done to extend self-supervised learning to different modalities~\cite{self_sup_audio_visual,self_sup_audio_visual2} or to audio representations only~\cite{self_sup_audio,Chorowsky_2019,cpc_deepmind,ravanelli2018learning}. With this regard, a recent trend consists of learning speech representations using a neural network encoder followed by a binary discriminator~\cite{cpc_deepmind,hjelm2018learning,ravanelli2018learning}. 

Despite recent progress, applying self-supervised learning to speech remains challenging. Speech signals are not only high-dimensional, long, and variable-length sequences, but also entail a complex hierarchical structure that is difficult to infer without supervision (phonemes, syllables, words, etc.). It is thus hard to find a single self-supervised task that can learn general and meaningful representations able to capture this latent structure.

To mitigate this issue, we propose to jointly tackle multiple self-supervised tasks using an ensemble of neural networks that cooperate to discover good speech representations. The intuition is that each self-supervised task may bring a different view or soft constraint on the learned representation. Even though not all the self-supervised tasks may help for the supervised problem of interest, there is likely a subset of them that could be useful. Another important implication is that our approach requires consensus across tasks, imposing several constraints into the learned representations. 
This way, our approach is more likely to learn general, robust, and transferable features, and less likely to focus on superficial features of the signal which may be sufficient for the given training data but are insufficient when considering broader types of data. 
To highlight the latter property, we call our proposed architecture the problem-agnostic speech encoder (PASE). 
PASE encodes the raw speech waveform into a representation that is fed to multiple regressors and discriminators. Regressors deal with standard features computed from the input waveform, resembling a decomposition of the signal at many levels. 
Discriminators deal with either positive or negative samples and are trained to separate them by minimizing binary cross-entropy~\cite{ravanelli2018learning}. 
Both regressors and discriminators (hereinafter called \textit{workers}) contribute to add prior knowledge into the encoder, which turns out to be crucial to derive meaningful and robust representations.

Our experiments suggest that PASE is able to discover robust representations from the raw speech waveform directly. We find that such representations outperform more traditional hand-crafted features in different speech classification tasks such as speaker identification, emotion classification, and automatic speech recognition. Interestingly, even though our representations are learned from a clean data set, the derived features turn out to work well also when processing speech that is corrupted by a considerable amount of noise and reverberation. PASE is designed to be efficient and fully parallelizable, and it can be seen as a first step towards a universal speech feature extractor. Moreover, PASE can be used as a pre-trained network avoiding to train models from scratch for each new task, as commonly done for computer vision models~\cite{vgg,inception}. 
PASE code and pre-trained model are available from \url{https://github.com/santi-pdp/pase}.

\begin{figure}[t]
    \centering
    \includegraphics[width=0.95\linewidth]{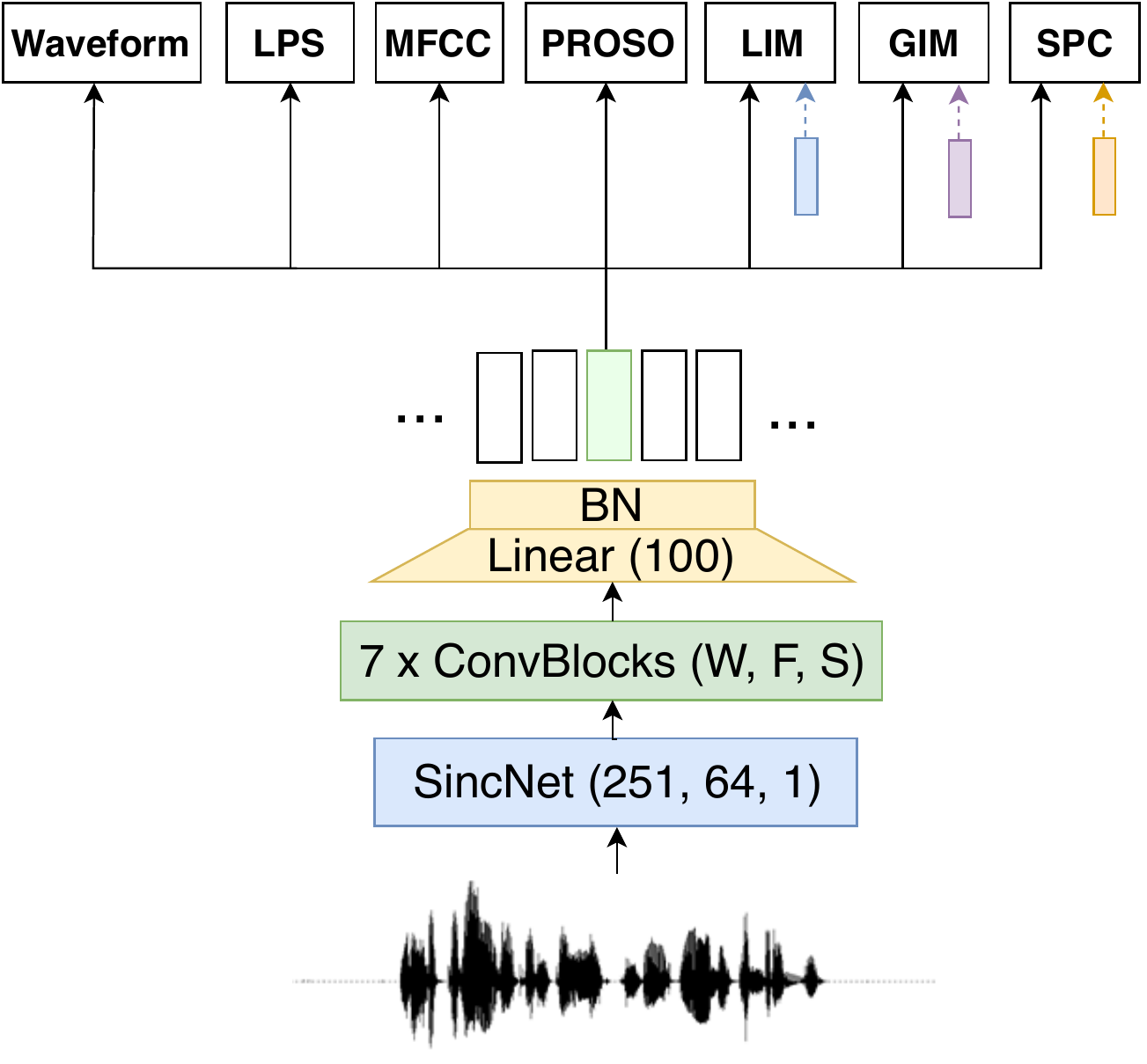}
    \caption{The PASE architecture, with the considered workers. 
    }
    \label{fig:PASE}
\end{figure}


\section{Problem-agnostic Speech Encoder}
The PASE architecture, depicted in Figure~\ref{fig:PASE}, is composed of a fully-convolutional speech encoder, followed by seven multilayer perceptron (MLP) workers, which cooperatively solve different self-supervised tasks. We now describe these modules.

\subsection{Encoder}
\label{sec:architecture}

The first layer of the encoder is based on the recently-proposed SincNet model~\cite{ravanelli2018speaker}. SincNet performs the convolution of the raw input waveform with a set of parameterized sinc functions that implement rectangular band-pass filters. 
An interesting property of SincNet is that the number of parameters does not increase with the kernel size. Similarly to~\cite{ravanelli2018speaker,sincnet_irasl}, we use a large kernel width $W=251$ to implement $F=64$ filters with a stride $S=1$. 
The subsequent layers are composed of a stack of 7~convolutional blocks (Fig.~\ref{fig:PASE}). Each block employs a one-dimensional convolution, followed by batch normalization (BN)~\cite{batchnorm}, and a multi-parametric rectified linear unit (PReLU) activation~\cite{he2015delving}. For the 7~blocks we use kernel widths $W=\{20, 11, 11, 11, 11, 11, 11\}$, $F=\{64, 128, 128, 256, 256, 512, 512\}$ filters, and strides $S=\{10, 2, 1, 2, 1, 2, 2\}$. An additional layer performs a convolution with $W=1$ that projects 512~features to embeddings of dimension 100. The final PASE representation is produced by a non-affine BN layer that normalizes by the mean and variance of each dimension.

Note that, similarly to common speech feature extractors based on the short-time Fourier transform, we emulate an overlapping sliding window using a set of convolutions. The convolution, in fact, 
employs a sliding kernel over the signal that extracts localized patterns at different time shifts. 
In our case, we use stride factors $S > 1$ for most of the convolutional blocks, such that the input signal is decimated in time by a factor of 160. Therefore, given an input waveform of $T$ samples, the amount of output feature vectors (frames) is $N = \frac{T}{160}$. At 16\,kHz, this is equivalent to a 10\,ms stride, similar to common speech processing pipelines. 
The receptive field of the encoder is about 150\,ms. 

\subsection{Workers}
\label{sec:unsupervised_training}

Workers are fed by the encoded representation and solve seven~self-supervised tasks, defined as regression or binary discrimination tasks (Fig.~\ref{fig:PASE}). In all cases, workers are based on very small feed-forward networks, composed of a single hidden layer of 256 units with PReLU activation (the only exception is the waveform worker, see below). Notice that we here employ simple networks on purpose. This way, we encourage the encoder, and not the workers, to discover high-level features that can be successfully exploited even by classifiers with limited capacity. 


We first consider the use of regression workers, which break down the signal components at many levels in an increasing order of abstraction. These workers are trained to minimize the mean squared error (MSE) between the target features and the network predictions (again the waveform worker is an exception, see below). Features are extracted with librosa~\cite{brian_mcfee_2019_2564164} and pysptk~\cite{pysptk} using default parameters, if not stated otherwise. As regression workers we consider:

\begin{itemize}
    \item \textbf{Waveform:} we predict the input waveform in an auto-encoder fashion. 
    The waveform decoder employs three deconvolutional blocks with strides 4, 4, and 10 that upsample the encoder representation by a factor of 160.  After that, an MLP of 256 PReLU units is used with a single output unit per time-step.
    This worker learns to reconstruct waveforms by means of mean absolute error (L1) minimization. The choice of L1 is driven by robustness, as the speech distribution is very peaky and zero-centered with prominent outliers~\cite{pascual2017segan}.
    \item \textbf{Log power spectrum (LPS):} as with the next features, we compute it using a Hamming window of 25\,ms and a step size of 10\,ms, with 1025~frequency bins per time step.
    \item \textbf{Mel-frequency cepstral coefficients (MFCC):} we extract 20~coefficients from 40 mel filter banks (FBANKs). 
    \item \textbf{Prosody:} we also predict four basic  features per frame, namely the interpolated logarithm of the fundamental frequency, voiced/unvoiced probability, zero-crossing rate, and energy. These features are called  ``Prosody'', inheriting a terminology often used in emotion recognition \cite{emotion1,emotion2}. 
\end{itemize}


Next, we also consider three binary discrimination tasks, learning a higher level of abstraction than that of signal features. These tasks rely on a pre-defined sampling strategy that draws an anchor $x_{a}$, a positive $x_{p}$, and a negative $x_{n}$ sample from the pool of PASE-encoded representations available in the training set. The reference anchor $x_{a}$ is an encoded feature extracted from a random sentence, while $x_{n}$ and $x_{p}$ are encodings drawn using the different sampling strategies described below. An MLP then minimizes the following formulation of the binary cross-entropy:
\begin{equation*}
\label{eq:ce}
L= \mathbb{E}_{X_p}[\log(g(x_a,x_p))] + \mathbb{E}_{X_n}[\log(1-g(x_a,x_{n}))],
\end{equation*}
where $g$ is the discriminator function, and $\mathbb{E}_{X_p}$ and $\mathbb{E}_{X_n}$ denote the expectation over positive and negative samples, respectively. Intuitively, by minimizing $L$, the model learns a speech embedding such that positive examples end up closer to their anchors than the corresponding negatives. Notice that the encoder and the discriminators are not adversarial here, but must cooperate to derive good representations.
In this work, we explore the following approaches to sample positive and negative examples:
\begin{itemize}
  \item \textbf{Local info max (LIM):} as proposed in~\cite{ravanelli2018learning}, we draw the positive sample from the same sentence of the anchor and a negative sample from another random sentence that likely belongs to a different speaker. Since the speaker identity is a reliable constant factor within random features of the same sentence, this worker can learn a representation that embeds this kind of information.
  
  \item \textbf{Global info max (GIM):} 
  in this and the subsequent worker, we compare global representations rather than local ones. 
  The anchor representation is obtained by averaging all the PASE-encoded frames of a random utterance within a long random chunk of 1\,s. The positive sample is similarly derived from another random chunk within the same sentence, while the negative one is obtained from another sentence. This way, we encourage the encoder to learn representations containing high-level information on the input sequence, that are hopefully complementary to those learned by LIM. GIM is also related to Deep InfoMax~\cite{hjelm2018learning}, which recently proposed to exploit local and global samples to learn image representations.

  \item \textbf{Sequence predicting coding (SPC):} in this case, the anchor is a single frame, while positive and negative samples are randomly extracted from its future and past elements. In particular, $x_{p}$ contains 5~consecutive future frames, while $x_n$ gathers 5~consecutive past ones. 
  To make the task less trivial, we avoid sampling inside the current-frame receptive field (150\,ms). On the other hand, to avoid making this task too complex or even unfeasible, we sample up to 500\,ms away from the anchor. We expect this worker to capture information about the sequential order of the frames and the signal causality, encouraging PASE to embed a longer time contextual information. This approach is similar to the sampling strategy used in the contrastive predicting coding work~\cite{van2017neural}. The main difference is that our negative sample is extracted from the past of the same sentence, rather than coming from a different one. 

\end{itemize}

\subsection{Self-supervised Training}
\label{sec:exp_setup_semisuptrain}

Encoder and workers are jointly trained with backpropagation by optimizing a total loss that is computed as the average of each worker cost.
Within the encoder, the gradients coming from the workers are thus averaged as well, and the optimization step will update its parameters pointing to a direction that is a compromise among all the worker losses~\cite{Serra18AIRD}.
To balance the contribution of each regression loss, we standardize all worker outputs using their mean and variance train set statistics, before computing the MSE.
The encoder and the workers are optimized with Adam~\cite{adam}, using an initial learning rate of $5\cdot 10^{-4}$ which is halved every 30 epochs. We use mini-batches of 32 waveform chunks, each with 16\,k samples corresponding to 1\,s at a 16\,kHz sampling rate. The system is trained for 150~epochs (i.e., until the validation losses reach a plateau for all the workers). 

\subsection{Usage in Supervised Classification Problems}
\label{sec:exp_setup_classif}

The representations discovered by the encoder can be later used for supervised classification in different ways. One possibility is to keep the encoder frozen while training the classifier (PASE-Frozen). The encoder is thus used as a standard feature extractor and the features do not dynamically change during training. A better way consists of fine-tuning both the encoder and classifier during supervised training (PASE-FineTuned). This way, the extracted features are further optimized to better adapt themselves to the application of interest. 
For comparison, our results also include the case where PASE is trained on the supervised task from scratch, with random initialization (PASE-Supervised).

\section{Corpora and Tasks}
\label{sec:exp_setup_corpora}

The self-supervised training of PASE is performed with the portion of the LibriSpeech dataset~\cite{librispeech} used in~\cite{ravanelli2018learning}. 
Speech sentences have been randomly selected to exploit about 15\,s of training material for each of the 2484~speakers. 

To assess the quality of the learned representations, we consider three supervised problems: (1) speaker identification (Speaker-ID), (2) speech emotion classification (Emotion), and (3) automatic speech recognition (ASR). 
For speaker identification, we use the VCTK dataset~\cite{veaux2016superseded}, which contains 109~speakers with different English accents. 
To make this task more challenging and realistic, we consider a subset of it that only contains 11\,s of training for each speaker.
For emotion recognition, we use the English utterances of the INTERFACE dataset~\cite{hozjan2002interface}.
This corresponds to approximately 3\,h for training, 40\,min for validation, and 30\,min for test. 
For speaker and emotion recognition, the neural posterior probabilities are averaged over all the time frames and we take the class with the highest score.
To evaluate the capability of PASE to learn phoneme representations, a first set of ASR experiments is performed with the standard TIMIT dataset~\cite{timit}. Next, to assess our approach in more challenging noisy and reverberant conditions, in Section~\ref{sec:res_transfer} we use the DIRHA dataset~\cite{dirha_asru}. 
Training and validation sets are based on the original WSJ-5k corpus (consisting of 7138~sentences uttered by 83~speakers) that is contaminated with a set of impulse responses  measured in a real apartment.  
The test set is composed of 409~WSJ sentences uttered by six American speakers and is based on real recordings in a domestic environment with a reverberation time of 0.7\,s and an average signal-to-noise ratio of about 10\,dB. ASR experiments are performed with the PyTorch-Kaldi toolkit~\cite{pytorch_kaldi} and are based on the DNN-HMM framework. The DNN is trained to predict context-dependent phones and an HMM decoder is later employed to retrieve the sequence of phonemes for TIMIT or words for DIRHA (using the language models of the Kaldi recipes~\cite{kaldi_short}). 

\section{Results}
\label{sec:res}

\subsection{Worker Ablation}

First of all, we study whether all considered workers contribute to the final accuracy of PASE, and assess their impact on different target problems. To do so, we retrain the encoder discarding one of the workers at a time. We then extract PASE features (using the frozen encoder described in section~\ref{sec:exp_setup_classif}), and we use them to feed MLP classifiers that solve the considered supervised problems. The experiments in this section are conducted with simple MLP classifiers based on a single layer, except for ASR, where we use three layers. 

The classification accuracies of Table~\ref{tab:ablation} show that no worker is dispensable. The best results are achieved with all workers, and we never observe performance improvements when discarding any of them. Nevertheless, while some workers are helpful for all the speech tasks, the benefits of some others turn out to be more application-dependent. For instance, Waveform, LPS, and MFCC regressors are generally helpful for all the applications, since they force the encoded representation to retain low-level information of the speech signal itself. The MFCC worker, in particular, is the most crucial one since it injects valuable prior knowledge on the most important frequency bands of the speech sequence. The prosody worker, instead, has a remarkable and expectable impact on emotion recognition only ($+$131\% in relative error). This is due to the fact that our prosody features are correlated with intonation, expressiveness, and voicing, which are crucial clues for detecting emotion. LIM and GIM seem to be more helpful for Speaker-ID and Emotion rather than for ASR. These workers are designed to extract high-level information of speech that can be better exploited by higher-level classification tasks. A similar trend is observed for the SPC worker. This tends to extract longer contextual information, which turns out to be helpful for speaker and emotion recognition ($+$16\% and in $+$13\% relative error, respectively). 
The adopted receptive field of 150\,ms, instead, embeds a context large enough for a DNN-HMM ASR system, as observed in~\cite{ravanelli_context}.

\begin{table}[t]
\centering
\caption{Accuracies using PASE and an MLP as classifier. Rows below the ``all workers'' model report absolute accuracy loss when discarding each worker for self-supervised training.}
\label{tab:ablation}
\setlength{\tabcolsep}{4pt}
\begin{tabular}{l|ccc}
    \hline 
     \multicolumn{1}{c|}{Model} & \multicolumn{3}{c}{Classification accuracy [\%]} \\
            & Speaker-ID & Emotion & ASR \\
            & \textit{(VCTK)} & \textit{(INTERFACE)} & \textit{(TIMIT)} \\
     \hline
     
     PASE (All workers)  & 97.5 & 88.3 &  81.1\\

     ~$-$ Waveform & $-$1.3 & $-$3.9 & $-$0.3 \\
     
    ~$-$ LPS & $-$1.5 & $-$5.3 & $-$0.5 \\
    
     ~$-$ MFCC & $-$2.4 & $-$3.2 & $-$0.7 \\
     
     ~$-$ Prosody & $-$0.5 & $-$5.3 & $-$0.1 \\

     ~$-$ LIM & $-$0.8 & $-$1.3 & $-$0.0 \\

     ~$-$ GIM & $-$0.6 & $-$0.5  & $-$0.3 \\

     ~$-$ SPC & $-$0.4 & $-$1.6 & $-$0.0 \\

     \hline
\end{tabular}
\end{table}

\subsection{Comparison with Standard Features}

We now compare our PASE representations with more standard features such as MFCCs and FBANK~\cite{brian_mcfee_2019_2564164}. Despite being proposed more than 40 years ago~\cite{Davis80-COP}, these coefficients are still the most common speech features, and it is not easy to find alternatives that consistently outperform them. To provide a more fair comparison, MFCCs and FBANK are gathered in context windows that embed contextual information of about 150\,ms (similar to the receptive field of the encoder). MFCCs are also augmented with their first and second derivatives. As mentioned, we also compare with the purely supervised version of PASE, trained from scratch on the target task.  

Table~\ref{tab:acc_final} shows the classification accuracies obtained with both MLP and recurrent neural network (RNN) classifiers based on gated recurrent units (GRU) \cite{gru}. 
The hyperparameters of all classifiers (number of hidden layers and neurons, learning rate, batch sizes, dropout rates, etc.) are independently tuned on the validation set and for each problem. 
PASE features provide most of the times a performance better than MFCCs and FBANKs, even when freezing the encoder (PASE-Frozen). 
The performance improvement becomes more evident when pre-training the encoder and fine-tuning it with the supervised task of interest (PASE-FineTuned). This approach consistently provides the best performance over all the tasks and classifiers considered here. 
Our best Speaker-ID result compares favorably with some recent works on the same dataset, such as~\cite{VCTK_paper} and~\cite{DilatedRNN}.
Interestingly, our best emotion recognition system achieves an accuracy of 97.7\%, which outperforms the human-level performance (80\%) measured in~\cite{hozjan2002interface} for the whole INTEFACE corpus.  
The phoneme accuracy of 85.3\% on the TIMIT dataset (an error rate of 14.7\%) is a competitive performance as well, especially when compared to state-of-the-art results that do not use complex techniques as system combination, speaker adaptation, or multiple steps of lattice rescoring and decoding~\cite{kaldi_short,pytorch_kaldi,li_gru,TIMIT_survey}.  

\begin{table}[t]
\centering
\caption{Accuracy comparison on the considered classification tasks using MLPs and RNNs as classifiers.}
\label{tab:acc_final}
\setlength{\tabcolsep}{3pt}
\begin{tabular}{l|cccccc}
    \hline 
    \multicolumn{1}{c|}{Model}  & \multicolumn{6}{c}{Classification accuracy [\%]} \\
            &  \multicolumn{2}{c}{Speaker-ID} & \multicolumn{2}{c}{Emotion} & \multicolumn{2}{c}{ASR}\\
           &  \multicolumn{2}{c}{\textit{(VCTK)}}  & \multicolumn{2}{c}{\textit{(INTERFACE)}}         & \multicolumn{2}{c}{\textit{(TIMIT)}}\\ \hline
            &  MLP & RNN & MLP & RNN & MLP & RNN \\
     \hline

     MFCC  & 96.9 & 72.3 & 90.8 & 91.1 & 81.1 & 84.8\\

     FBANK  & 98.4 & 75.1 & 94.1 & 92.8 & 80.9 & 85.1\\
          
     PASE-Supervised  & 97.0 & 80.5 & 93.8 & 92.8 & 82.1 & 84.7 \\
     
     \hline
     PASE-Frozen & 97.3 & 82.5 & 91.5 & 92.8 & 81.4 & 84.7\\
     
     PASE-FineTuned & \textbf{99.3} & \textbf{97.2}  & \textbf{97.7} & \textbf{97.0} & \textbf{82.9} &  \textbf{85.3}\\
     
     \hline
\end{tabular}
\end{table}

\subsection{Transferability}
\label{sec:res_transfer}

Finally, we study the exportability of PASE to acoustic conditions that are very different from the clean one used to train it. 
Table~\ref{tab:dirha} reports the results obtained with the DIRHA dataset, which contains speech signals characterized by considerable noise and reverberation. We here employ the same version of PASE encoder used so far (trained on clean LibriSpeech data) coupled with a GRU classifier.
Interestingly, PASE clearly outperforms the other systems. Even the frozen version of PASE overtakes FBANKs, MFCCs, and the supervised training baseline. PASE-FineTuned also outperforms our previous results obtained with the standard SincNet model~\cite{sincnet_irasl}. This result suggests the ability of PASE to effectively transfer its representation abstractions to different acoustic scenarios. 

\begin{table}[t]
\centering
\caption{Word error rate (WER) obtained on the DIRHA corpus.}
\label{tab:dirha}
\setlength{\tabcolsep}{10pt}
\begin{tabular}{l|c}
    \hline 
            & WER [\%] \\
     \hline
     MFCC  & 35.8  \\
     FBANK  & 34.0  \\
     PASE-Supervised  & 33.5 \\ 
     \hline
     PASE-Frozen & 32.5 \\
     PASE-FineTuned & \textbf{29.8} \\

     \hline
\end{tabular}
\end{table}


\section{Conclusion} 
\label{sec:conc}
The proposal of this work was twofold. On the one hand, we proposed a multi-task self-supervised approach to learn speech representations. On the other hand, we provided an effective and exportable speech encoder that conveys waveforms into a sequence of latent embeddings. As evidenced by the considered problems, the discovered embeddings turn out to carry important information of the speech signal, related to, at least, speaker-identity, phonemes, and emotional cues. Learnt embeddings also showed their potential for of transferability to different datasets, tasks, and acoustic conditions.
Moreover, PASE is easily extendable as a semi-supervised framework and can embed in the future many other self-supervised tasks.

\section{Acknowledgements}

This research was partially supported by the project TEC2015-69266-P (MINECO/FEDER, UE), Calcul Qu\'ebec, and Compute Canada. 
We also thank Loren Lugosch, Titouan Parcollet, and Maurizio Omologo for helpful comments. 


\bibliographystyle{IEEEtran}

\bibliography{mybib}


\end{document}